\documentclass{article}
\usepackage{ijcai21}

\usepackage{times}
\usepackage{soul}
\usepackage{url}
\usepackage[hidelinks]{hyperref}
\usepackage[utf8]{inputenc}
\usepackage[small]{caption}
\usepackage{graphicx}
\usepackage{booktabs}
\usepackage{amsmath}
\usepackage{amsthm}
\usepackage{amssymb}
\usepackage{amsfonts}
\usepackage{booktabs}
\usepackage{algorithm}
\usepackage{algorithmic}
\usepackage{multirow}
\usepackage{subfigure}
\newtheorem{problem}{Problem}

\title{TrafficStream: A Streaming Traffic Flow Forecasting Framework \\
Based on Graph Neural Networks and Continual Learning}

\author{
Xu Chen\footnote{Corresponding Author}\and
Junshan Wang\and
Kunqing Xie\\
\affiliations
School of Electronics Engineering and Computer Science, Peking University, Beijing, China\\
\emails
\{sylover,wangjunshan,kunqing\}@pku.edu.cn
}

\begin{document}

\maketitle

\begin{abstract}

With the rapid growth of traffic sensors deployed, a massive amount of traffic flow data are collected, revealing the long-term evolution of traffic flows and the gradual expansion of traffic networks. How to accurately forecasting these traffic flow attracts the attention of researchers as it is of great significance for improving the efficiency of transportation systems. 
However, existing methods mainly focus on the spatial-temporal correlation of static networks, leaving the problem of efficiently learning models on networks with expansion and evolving patterns less studied. 
To tackle this problem, we propose a \textit{Streaming Traffic Flow Forecasting Framework}, \textbf{TrafficStream},  based on \textit{Graph Neural Networks (GNNs)} and \textit{Continual Learning (CL)}, achieving accurate predictions and high efficiency. 
Firstly, we design a traffic pattern fusion method, cleverly integrating the new patterns that emerged during the long-term period into the model. A JS-divergence-based algorithm is proposed to mine new traffic patterns. 
Secondly, we introduce CL to consolidate the knowledge learned previously and transfer them to the current model. Specifically, we adopt two strategies: historical data replay and parameter smoothing.
We construct a streaming traffic dataset to verify the efficiency and effectiveness of our model. Extensive experiments demonstrate its excellent potential to extract traffic patterns with high efficiency on long-term streaming network scene. The source code is available at https://github.com/AprLie/TrafficStream.

\end{abstract}
\section{Introduction}
In recent years, the development of the economy and society has prompted the construction of transportation facilities like roads, toll stations, subway stations, etc. 
To monitor the real-time traffic status of these facilities, the number of traffic sensors deployed also gains rapid growth.
As a result, more high-quality traffic flow data is recorded, revealing the expansion of traffic networks and the evolution of traffic flows.

Traffic flow forecasting is one of the most crucial tasks of Intelligent Transportation System (ITS), which significantly affect our daily routines. Nevertheless, the intricate spatial-temporal correlation of traffic flow raises the difficulty of accurate forecasting. To tackle this problem, researchers introduce various cutting-edge methods into this field, such as CNNs \cite{zhang2017deep}, LSTMs \cite{dcrnn} and Transformers \cite{wang2020traffic}.
Impressed by the promising performance of GNNs on graph structure data, researchers are devoted to integrating GNNs into traffic flow forecasting model for mining dependency on spatial-temporal aspect, which becomes a great success \cite{stgcn,stsgcn,tssrgcn}.
However, along with the improvement of model performance, there comes a considerable increase in the number of parameters. Besides, they still face the challenges of long-term streaming traffic networks.

The long-term streaming networks, referring to traffic networks with changing flows of nodes and expanding structures in a long-term period, i.e., several years, have been less studied, whereas the real-world scene is like this. The simplest way is to retrain a model every year. However, the massive amount of parameters of a new model are severe challenges to computing resources and the amount of data, especially when datacenters store data in a queue-like manner. Besides, it will be difficult to generate accurate predictions via less data from new added stations. 
Considering that previous models have been well trained and for resource conservation, it is natural to transfer their knowledge to the model in the current year.

Directly using previous models still needs to address two critical problems. 
Firstly, previous models should accommodate the new data. Unlike short-term data whose trends are more stable and regular, traffic flow patterns gradually evolve in the long-term streaming networks. The inconsistency between patterns of new data and historical data will reduce the accuracy of forecasting. Besides, due to network expansion, the topological structures of traffic networks corresponding to the new data are also different from the previous ones. So the spatial dependency captured by previous models may not be suitable anymore. 
Secondly, knowledge extracted by previous models should be consolidated as there is rich information about previously learned patterns \footnote{Previously learned patterns include long-term stable patterns and temporarily disappearing patterns.}. Therefore, \textit{how to efficiently capture patterns from new data while consolidating historical knowledge} becomes our motivation.

For efficiency and effectiveness, we propose a Streaming Traffic Flow Forecasting Framework based on GNNs and Continual Learning, TrafficStream. 
On the one hand, we design a traffic pattern fusion approach to integrate the unknown patterns of new nodes and changing patterns of existing nodes into the model. To discover these new patterns, we propose an algorithm based on JS-divergence to measure the changes of node features depending on the model. 
On the other hand, in order to consolidate historical traffic knowledge and transfer it to the current model for better prediction, we adopt strategies of historical information replay and parameter smoothing from data and model perspectives, respectively. 
Finally, a long-term streaming network flow dataset is constructed to verify the performance of TrafficStream.

Our main contributions are summarized as follows:
\begin{itemize}
    \item We explore the problem of streaming traffic flow forecasting based on GNNs. To our best knowledge, it is the first time that GNNs are introduced in the scenario of changing flows of nodes and expanding network structures in long-term streaming traffic networks.
    \item We propose a novel unified framework for GNN-based models to address the difficulty of fusing historical knowledge and new flow data. We design different components for training models on expanding networks and nodes with new evolutionary patterns. Information replay and knowledge smoothing components are adopted to consolidate historical knowledge.
    \item To verify the effectiveness and efficiency of our framework, we construct a long-term streaming network dataset, PEMSD3-Stream, based on real-world traffic data. The performance indicates the superiority of our model in consuming time with comparative accuracy.
\end{itemize}
\section{Related Work}
\subsection{Traffic Flow Forecasting}
Recent years have witnessed the rapid development of ITS, attracting researchers to solve one of the key problems, traffic flow forecasting. 
Statistical models, such as VAR \cite{var} and SVR \cite{svr} simplify the forecasting into individual time-series prediction.
These methods failed to capture the spatial-temporal correlation between different locations. In addition, training such a model for all locations results in poor performance, while each location with a model will increase cost in both consuming time and storage.
Along with the development of deep learning, researchers apply various frameworks to overcome the difficulty of understanding the pattern of spatial-temporal data. 
DMVST-Net \cite{yao2018b} integrates CNN and long-short term memory (LSTM) to extract spatial and temporal dependencies. 
STGCN \cite{stgcn} introduces graph convolution for mining spatial patterns and CNN with the gated mechanism for temporal features.
Graph WaveNet \cite{graphwavenet} is proposed to adaptively learn the spatial adjacency matrix and expand the receptive field along the time axis.
STSGCN \cite{stsgcn} simultaneously models the spatial-temporal correlation via a localized spatial-temporal graph.
Although these methods achieve impressive results on traffic flow forecasting task, they are not designed for long term streaming network scene.

\subsection{Continual Learning on Graphs}

Continual learning, also called as life-long learning and incremental learning, is a technique to train the model sequentially when data from different tasks arrives in a streaming manner. Generally, existing methods can be divided into three categories: regularization-based methods \cite{ewc}, replay-based methods \cite{experience_replay} and parameter isolation methods \cite{rusu2016progressive}, and all of them aim to learn knowledge from new tasks without forgetting knowledge from previously trained tasks. 
Traditional continual learning methods mainly focus on image processing, assuming that samples are independent to each other.

Recently, continual learning has been applied in the field of graph learning, which can deal with the efficient update of graph models, especially on large-scale dynamic graphs.
ContinualGNN \cite{wang2020streaming} proposes an incremental GNN model on streaming graphs by detecting new coming patterns efficiently and consolidating existing patterns via continual learning. Besides, ER-GNN \cite{zhou2020continual} and Feature Graph \cite{wang2020lifelong} presents different experience replay strategies for continual graph representation learning. 
However, these methods only consider the incremental learning of GNN models when the graph structures change, and thus can only be applied to scenarios such as social networks and knowledge graphs. They fail to handle the effective update of the spatial-temporal model in traffic networks when the temporal patterns of stations and the topological structures of networks evolve simultaneously.

\section{Preliminaries}
The long term streaming traffic network is represented as $G=(G_1,G_2,\dots,G_{\mathcal{T}})$\footnote{In this paper, we use $\tau \in\{1,2,\dots,\mathcal{T}\}$ to represent a long time interval that the traffic network structure remains unchanged within $\tau$. $t\in\{1,2,\dots,T\}$ denotes the time-steps traffic data is observed.} , where $G_{\tau} = G_{\tau-1}+\Delta G_{\tau}$ for all time intervals $\tau$, meaning the network keeps expanding over time. 
 Traffic network of the $\tau$-th year can be denoted as a graph $G_{\tau}=(V_{\tau},E_{\tau},A_{\tau})$, where $V_{\tau}$ is a finite node set with $|V_{\tau}|=N_{\tau}$ nodes. Nodes on the graph can represent sensors or stations. $E_{\tau}$ represents the edge set with edges connecting nodes from $V_{\tau}$ and $A_{\tau} \in \mathbb{R}^{N_{\tau}\times N_{\tau}}$ is the adjacency matrix. Devices deployed at nodes will periodically measure and record traffic status features, including flows, speed and occupancy. Flow data of all nodes on graph $G_{\tau}$ observed at time-step $t$ can be denote as $X^{t}_{\tau} \in \mathbb{R}^{N_{\tau}\times F}$, where $F$ is the number of traffic status features. 

The purpose of traffic flow forecasting is to provide accurate predictions of future traffic flow when historical data is given, which can be formulated as:
\begin{problem}
Given the long-term streaming traffic network $G=(G_1,G_2,\dots,G_{\mathcal{T}})$ with  $G_{\tau}=(V_{\tau}, E_{\tau}, A_{\tau})$ and the corresponding historical data $X_{\tau}=(X^{1}_{\tau},\ldots,X^{T}_{\tau})$, we are expected to learn a series of function $\Psi=(\Psi_{1},\Psi_{2},\dots,\Psi_{\mathcal{T}})$ for prediction of traffic flow series $\hat{Y_{\tau}}=\Psi_{\tau}(X_{\tau})=(\hat{Y}^{T+1}_{\tau},\ldots,\hat{Y}^{T+K}_{\tau})$ for all $N_{\tau}$ nodes in the next $K$ time-steps after time-step $T$, i.e.,:
\begin{equation}
    \Psi^*_{i} = \arg\min_{\Psi_{i}} ||\Psi_{\tau}(X_{\tau}) - Y_{\tau}||^2,
\end{equation}
where $Y_{\tau}^T\in \mathbb{R}^{N_{\tau}}$ denotes the ground-truth of the traffic flow at time-step $T$ and $Y_{\tau}=(Y^{T+1}_{\tau},\dots,Y^{T+K}_{\tau})$ represents the time series to be forecast.
\end{problem}

\section{TrafficStream}
In this section, we elaborate on the proposed framework of streaming traffic flow forecasting based on \textit{GNNs} and \textit{Continual Learning}, \textbf{TrafficStream}. 
We firstly introduce a simple GNN model as a surrogate model of those complex traffic flow forecasting methods. 
Our framework balances between learning new traffic patterns from new data and maintaining the historical knowledge learnt from previous data. 
New nodes with its 2-hops neighbors are used to construct a sub-graph for mining the influence of network expansion. Existing nodes whose traffic patterns change significantly are detected through an algorithm based on JS-divergence. To consolidate the previous knowledge, historical nodes of traffic networks are replayed and weighted smoothing constraints are imposed on the current training model. 
The details of TrafficStream are described in the following.

\subsection{Surrogate Traffic Flow Forecasting Model}
\label{sec:41}

The great success of graph neural networks (GNNs) have attracted the attention of researchers and they are introduced to capture spatial and temporal features of traffic flow data. 
Consequently, current studies on traffic flow forecasting are mainly based on GNNs: they are leveraged to extract spatial dependency between nodes together with sequence models like CNNs and RNNs mining temporal trends of traffic flow \cite{stgcn,graphwavenet,astgcn}. Recent studies simultaneously capture spatial-temporal correlation through applying GNNs on spatial-temporal fusion graphs constructed by data \cite{stsgcn,mengzhang2020spatialstfgnn}. 

To intuitively describe our framework, we follow their common design and propose a simple surrogate traffic flow forecasting model (\textbf{SurModel}) to represent those complex ones deployed in ITS. SurModel is composed of two GNN layers with a CNN layer sandwiched in between. Denote the input of $l$-th GNN layer in $\tau$-th year as $H^{l}_{\tau}\in\mathbb{R}^{N_{\tau}\times C_{\tau}^l}$ ($H^{0}_i=X_{\tau}$), spatial information obtained by the GNN layer can be formulated as: 
\begin{equation}
\label{eq:gcn}
    H^{l+1}_\tau = \sigma(A_{\tau}H^{l}_{\tau}W^l_{\tau 1}+IH^{l}_{\tau}W^l_{\tau2})
\end{equation}
where $W^l_{\tau1},W^l_{\tau2}\in\mathbb{R}^{C_{\tau}^l\times C_{\tau}^{l+1}}$ are learnable parameters. 
$I\in\mathbb{R}^{N_{\tau}\times N_{\tau}}$ is identity matrix and $\sigma$ is the activation function. 1D CNN is adopted to capture temporal patterns on the output of the first GNN layer and generate features for the second GNN layer. We apply skip connection to concatenate input $X_{\tau}$ with output of the second GNN layer $H^{2}_{\tau}$. The result will be fed into a fully connected layer to generate traffic flow in the next $K$ time-steps. SurModel is set to minimize the L2-loss between predicted value  $\hat{Y}_{\tau}$ and ground-true value $Y_{\tau}$:
\begin{equation}
\label{eq:loss}
\mathcal{L}(\hat{Y}_{\tau},Y_{\tau}) = \| \hat{Y}_{\tau} - Y_{\tau}\|^2.
\end{equation}

\subsection{New Traffic Pattern Fusion}

Traffic flow patterns are believed to be stable without accidents on the roads. Researchers tend to categorize them into daily pattern, weekly pattern, monthly pattern and so on. Based on this assumption, they are devoted to decomposing traffic flow into different patterns for deeper comprehension of traffic systems and better prediction of traffic status.
However, this assumption becomes invalid if we observe the flow in a long term: trends may evolves sluggishly over time and it is also possible that the network structures change due to deployment of new sensors or stations. The extracted patterns are not more accurate enough to describe traffic laws of the current network. 
To efficiently update these patterns under such situations, we propose two components for training on expanding networks and better extracting evolving patterns.

\subsubsection{Expanding Network Training}
Current GNN based traffic flow forecasting models have a huge amount of parameters, and the time complexities are $\mathcal{O}(N^2_{\tau-1})$ which can be attributed to matrix multiplication of GNN layers in Eq. \eqref{eq:gcn}. When network scale changes, retraining the model definitely raises the complexity to $\mathcal{O}(N_{\tau}^2)=\mathcal{O}((N_{\tau-1}+\Delta N_{\tau})^2)$. Here $\Delta N_{\tau}$ is the node set that only contains newly added nodes. In real-world traffic network, it is often the case that previous networks are sufficiently large while the newly deployed nodes only account for a small part. Thus, retraining seems more time consuming as $\Delta N_{\tau}\ll N_{\tau}$. Motivated by the scale between $\Delta N_{\tau}$ and $N_{\tau}$ as well as online learning methods, we firstly use $\Delta N_{\tau}$ to update the trained SurModel $\Psi_{\tau-1}$ as $\Psi_{\tau}$. Note that GNN layers aim to capture the dependency between nodes and its neighborhoods, we generate a sub-graph by selecting neighbor nodes of newly added nodes on $G_{\tau}$ within 2-hops. The complexity is reduced to $\mathcal{O}((\Delta N_{\tau}d^2)^2)$ where $d$ is the average degree of nodes.
Remember that on a traffic network nodes are connected to 2 nodes on a straight road and 4 at a crossroads, so that $d$ is around 2$\sim$4 and $\mathcal{O}((\Delta N_{\tau}d^2)^2)$ is much smaller than $\mathcal{O}(N_{\tau}^2)$.

\subsubsection{Evolution Pattern Detection}
In addition to the effect of network expansion, traffic patterns would not remain constant: people may change residences and workplaces or there may be some changes at the point of interest (POI), and thus travelling plans vary over time. It is necessary to detect nodes where current patterns sharply conflict with previous ones. We design a node detection algorithm to discover these nodes: given the last graph $G_{\tau-1}$ and current graph $G_{\tau}$, for all the nodes on $G_{\tau-1}$, we select flow data of the last week at $\tau-1$ and $\tau$. Denote these data as $X_{\tau-1}$ and $X_{\tau}$, and previous SurModel $\Psi_{\tau-1}$ without the last fully connected layer is $\psi_{\tau-1}$. Patterns captured from data can be represented as the output feature $U$ of $\psi_{\tau-1}$:
\begin{equation}
 U_{\tau} = \psi_{\tau-1}(X_{\tau}), \quad U_{\tau-1} = \psi_{\tau-1}(X_{\tau-1})
\end{equation}

To tackle the problem that features are different between weekdays and weekends, in the time dimension, we normalize $U_{\tau}$ and $U_{\tau-1}$ and utilize histogram of the normalize features as its distribution. Let $P_{\tau}$ and $P_{\tau-1}$ represent these two distribution of feature value. We aim to discover the similarity between these two distribution as it reveals the change of current and previous patterns.
Thus, Jensen–Shannon divergence (JSD or JS divergence) is introduced as the measurement of similarity:
\begin{equation}
	JSD(P_{\tau}||P_{\tau-1}) = \frac{1}{2}D(P_{\tau}||\bar{P})+\frac{1}{2}D(P_{\tau-1}||\bar{P})
\end{equation}
where $D(P||Q)=\sum_{x\in\mathcal{X}}P(x)\log (P(x)/ Q(x))$ is the Kullback–Leibler divergence and $\bar{P}=(P_{\tau}+P_{\tau-1})/2$.

If JSD of a node is pretty high, its pattern can be considered as evolving substantially. Finally, we choose the top 5$\%$ of nodes with the highest JSD score as candidate nodes for learning the new patterns. Similar to those newly added nodes, we also construct a sub-graph for GNN layers.

\subsection{Historical Traffic Knowledge Consolidation}
When models capture the new patterns through components mentioned above, another problem arises: the model may forget previous knowledge it has learnt because it focus on those new patterns only. The catastrophic forgetting phenomenon arouses us to consolidate historical traffic knowledge. One way to preserve existing patterns is to track back to the source where these knowledge from, i.e., historical flow data. The other is to force current training model $\Psi_{\tau}$ to remember useful knowledge from $\Psi_{\tau-1}$. Motivated by these two ideas, we design two components to consolidate knowledge from data and model perspectives. Details are presented as follows.

\subsubsection{Information Replay}

In order to maintain existing knowledge, we can train the model $\Psi_{\tau}$ with sampled historical information and make a replay. A simple strategy is to randomly sample flow data of some nodes, whereas it has large errors for these samples may be less representative. Rethinking the node detection algorithm based on JS divergence, nodes whose JSD scores are lower reveal that their flow is more stable, so that it is reasonable to choose these nodes. In the end, whose nodes with the lowest JSD scores are selected and construct a sub-graph to train $\Psi_{\tau
}$ as a replay.

\subsubsection{Parameter Smoothing}
For better preserving previous knowledge, we utilize the last model $\Psi_{\tau-1}$ to impose smoothing constraints on parameters of the current one $\Psi_{\tau}$, so as to avoid large deviation of the model and the occurrence of the forgetting problem. Adopting l2 constraints is a crude approach, where excessive restrictions on the model may make it lack the ability to learn new patterns. If we know which parameters are essential to the existing patterns of traffic flows and only restrict these parameters, we can get a more expressive model for forecasting.

Therefore, we introduce elastic weight consolidation (EWC) \cite{ewc}, adding a more flexible and exquisite smoothing term to the loss function: 
\begin{equation}
    \mathcal{L}_{smooth} = \lambda \sum_i \mathbf{F}_i (\Psi_{\tau}(i) - \Psi_{\tau-1}(i))^2,
\label{equ:smooth}
\end{equation}
where $\lambda$ is the weight of the smoothing term and $\mathbf{F}_i$ is the importance of the $i$-th parameter in the model $\Psi_{\tau-1}$, which is estimated using Fisher Information as
\begin{equation}
\mathbf{F}  = \frac{1}{|X_{\tau-1}|} \sum_{x \in X_{\tau-1}}[g(\Psi_{\tau-1};x) g(\Psi_{\tau-1};x)^T],
\end{equation}
where $g$ is the is first order derivatives of the loss.

Through the weighted smoothing term, the weight of the less essential parameters for historical data is smaller, and these parameters can better adapt to the new patterns. On the contrary, the more important parameters for historical data are more weighted, and thus their changes are restricted, ensuring the preservation of historical knowledge.

\section{Experiments}

\subsection{Dataset}
To verify the effectiveness and efficiency of our framework, we conduct some experiments on a real-world dataset, PEMS3-Stream.
PEMS3-Stream is collected by California Transportation Agencies (CalTrans) Performance Measurement System (PeMS) \cite{pems} in real-time by every 30 seconds. 
The data is aggregated into every 5-minutes interval from 30-seconds data instances. PEMS3-Stream contains traffic flow data in North Central Area from 2011 to 2017. We select data from July 10th to August 9th every year. The details are shown in Table \ref{tab:my-table}.

\begin{table}[b]
\centering
\resizebox{0.48\textwidth}{!}{%
\begin{tabular}{c|ccccccc}
\toprule
Year &  2011 &  2012 &  2013 &  2014 &  2015 &  2016 &  2017 \\ 
\midrule
\# Nodes &  655 &  715 &  786 &  822 &  834 &  850 &  871 \\ 
\midrule
\# Edges &  1577&  1929&  2316&  2536&  2594&  2691&  2788\\ 
\bottomrule
\end{tabular}
}
\caption{The statistic of PEMSD3-Stream dataset.}
\label{tab:my-table}
\end{table}

\subsection{Preprocessing}
To construct a high-quality dataset, we select sensors by the following rules: (1) the location shifting of a sensor in corresponding metadata should be less than 100 meters; (2) the missing ratio of data of a sensor should be less than 15$\%$; (3) the traffic network of PEMS3-Stream keeps expanding, meaning sensor $i$ at the $\tau$-th year must appear at the $(\tau+\Delta \tau)$-th year, $\Delta\tau\in\{1,2,\ldots, 7-\tau\}$.

The adjacency matrix $A_{\tau}$ at the $\tau$-th year can be extracted from metadata of sensors:
\begin{equation}
\begin{split}
&A_{\tau}[mn] = \left\{
\begin{aligned}
& \exp\left(-\frac{d_{mn}^2}{\sigma_d^2}\right)&&, \mbox{$ m \ne n$ and 
$d_{mn}\leq\epsilon$}\\
& 0&&, \mbox{otherwise} \\
\end{aligned}\right. 
\end{split}
\end{equation}
where $d_{ij}$ denotes the distance between sensor $i$ and $j$. $\sigma_d$ is set to 10 and $\epsilon$ is set to 1.0 to control the sparsity of $A_{\tau}$.

\begin{table*}[h]
\centering
\footnotesize
\setlength{\tabcolsep}{4pt}
\begin{tabular}{l|l|ccc|ccc|ccc|cc}
\toprule
\multicolumn{2}{c|}{PEMSD3}    & \multicolumn{3}{c|}{15 min} & \multicolumn{3}{c|}{30 min} & \multicolumn{3}{c|}{60 min} & \multicolumn{2}{c}{Running Time} \\ 
\midrule
\multicolumn{2}{c|}{Model}     & MAE     & RMSE    & MAPE    & MAE     & RMSE    & MAPE   & MAE     & RMSE    & MAPE & Total & Avg. (s) \\ 
\midrule
\multirow{5}{*}{Retrained} & SVR & 14.76 & 25.06 & 20.16 & 16.15 & 27.36 & 21.73 & 19.02 & 32.24 & 25.13&593.42&- \\
& GRU & 13.38 &	23.04 &	18.70 &	14.61 &	24.94 &	20.55 &	17.21 &	28.92 &	24.92 & 256.22 	& 7.96  \\
& STGCN & \textbf{12.30} &	\textbf{20.37} & \textbf{16.96}	&	14.29 &	23.69 & 20.69	&	17.45 &	28.53  & 25.77 & 6518.70 &	147.78  \\
& STSGCN & 13.02   & 20.99   & 17.34  & \textbf{13.62}   & \textbf{22.06}  & \textbf{17.97}  & \textbf{14.76}   & \textbf{24.03}   & \textbf{19.23}  & 12076.21           & 162.47 \\
& Retrained-STModel & 12.76 & 20.59 & 17.83 & 13.92 & 22.76 & 19.90 & 16.33 & 26.90 & 24.99 & 1232.58 & 32.45 \\
\midrule
\multirow{3}{*}{Online} & Static-STModel & 13.98 & 21.88 & 29.36 & 15.12 & 23.98 & 31.67 & 17.46 & 28.01 & 36.44 & 319.39  & 3.85  \\
& Expansible-STModel & 13.51 & 21.83 & 18.19 & 15.05 & 24.72 & 19.88 & 18.24 & 30.49 & 23.82 & 418.46  & 9.13  \\
\cmidrule{2-13}
& \textbf{TrafficStream} & \textbf{12.79} & \textbf{20.71} & \textbf{17.05} & \textbf{13.97} & \textbf{22.90} & \textbf{18.74} & \textbf{16.37} & \textbf{27.04} & \textbf{22.94} & 461.24  & 14.49 \\ 
\bottomrule
\end{tabular}
\caption{Overall results of traffic forecasting on PEMS3-Stream dataset.}
\label{tab:res}
\end{table*}

\begin{figure*}[!h]
\centering
\subfigure[15min MAE]{
    \includegraphics[width=0.66\columnwidth]{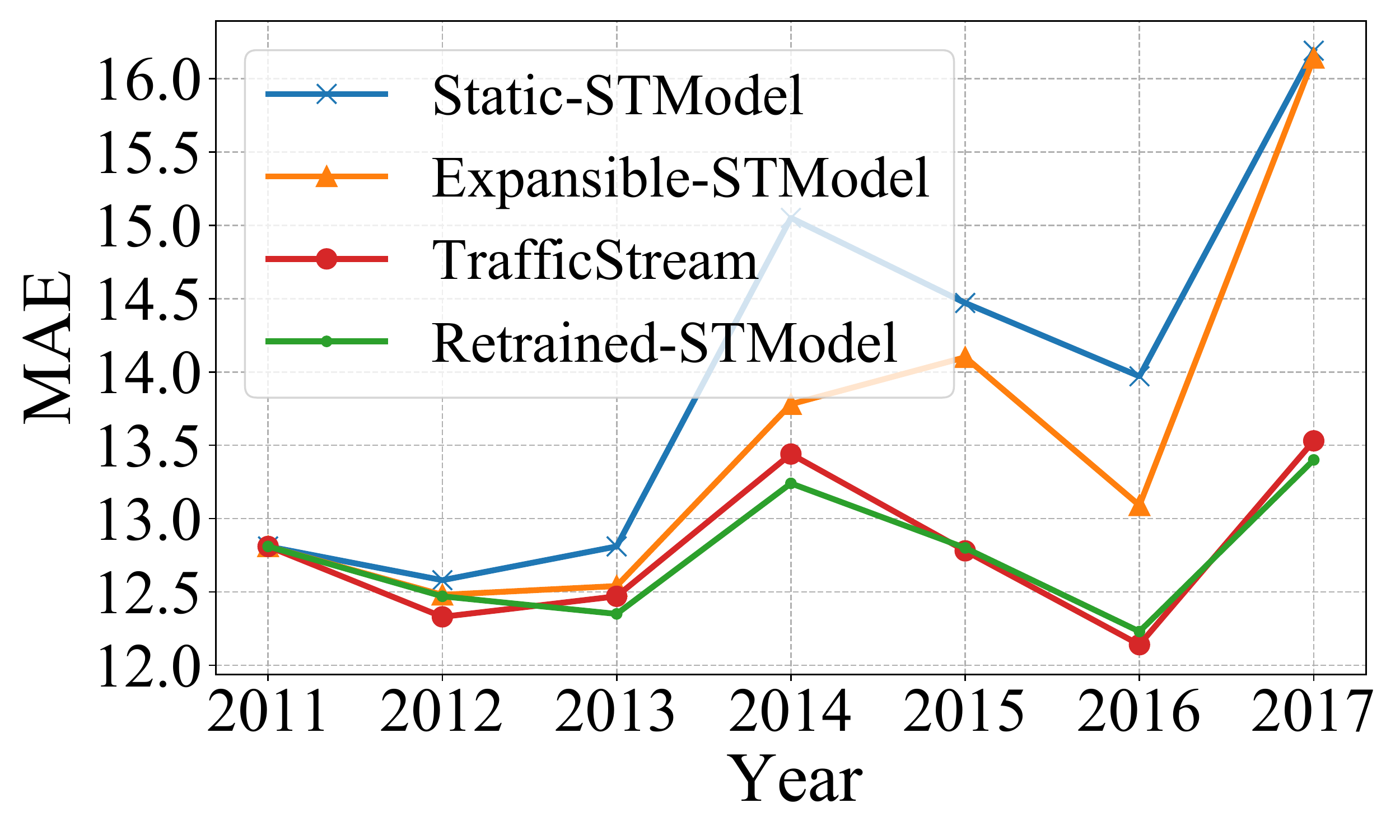}
    \label{fig:a}
}
\subfigure[15min RMSE]{
    \includegraphics[width=0.66\columnwidth]{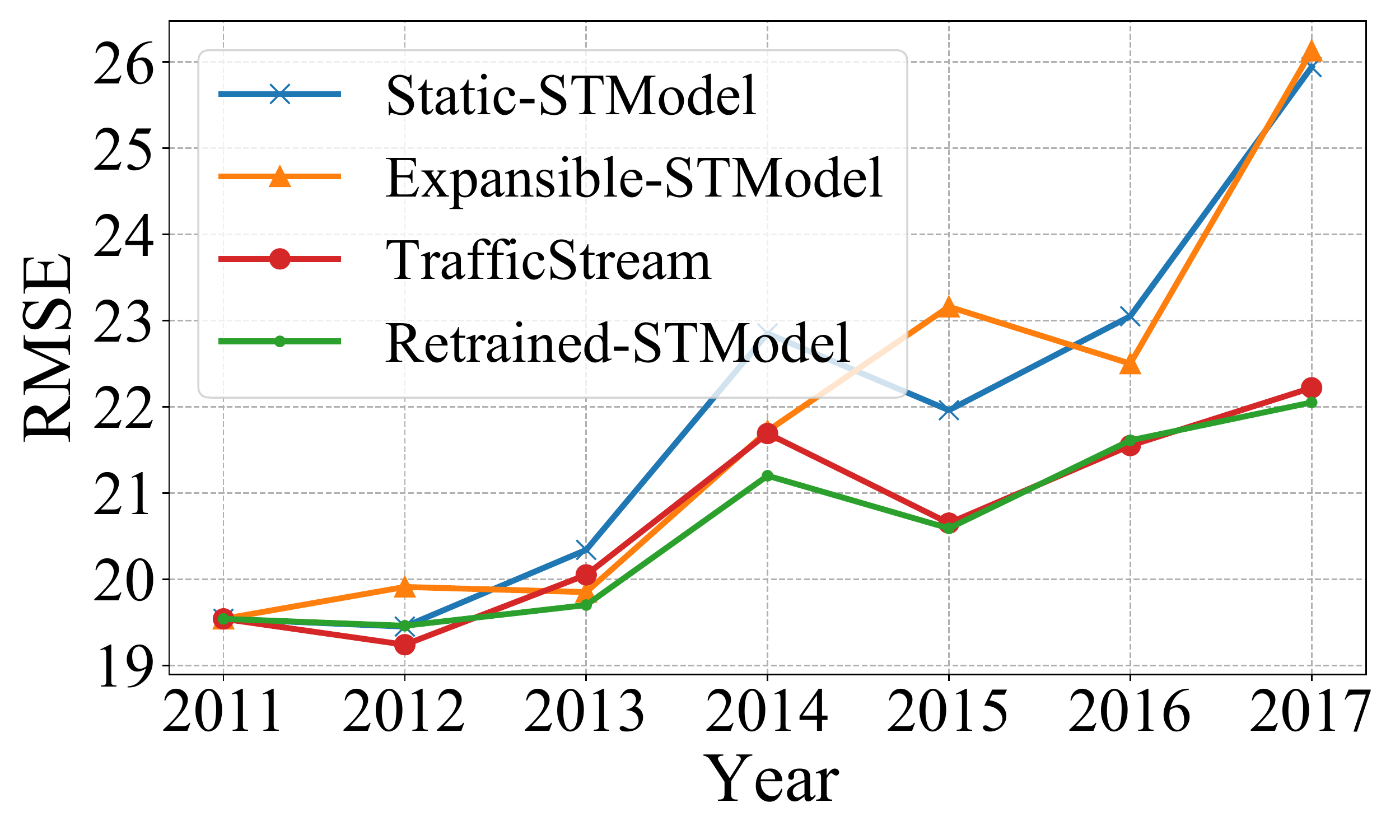}
    \label{fig:b}
}
\subfigure[Total running time]{
    \includegraphics[width=0.66\columnwidth]{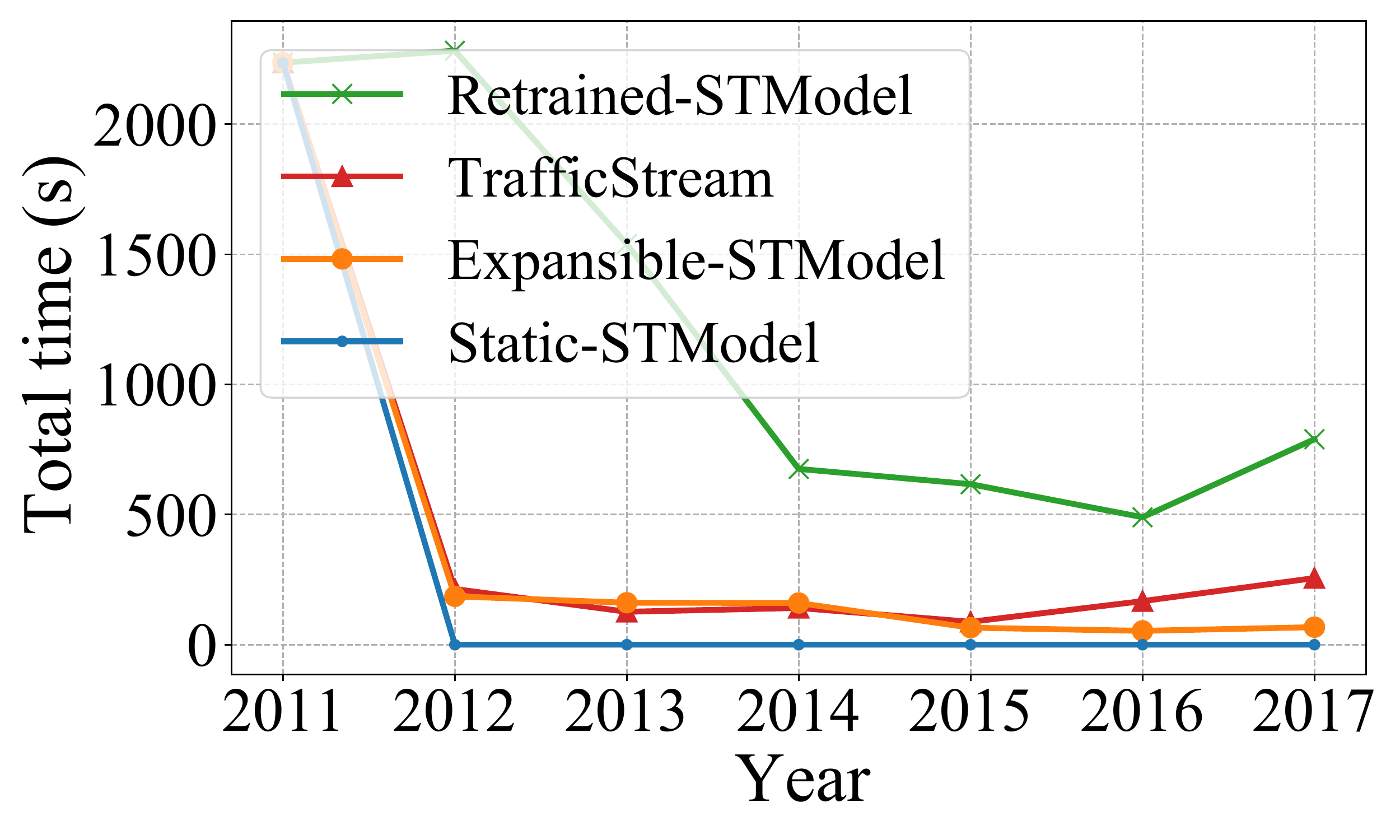}    
    \label{fig:c}
}
\subfigure[30min MAE]{
    \includegraphics[width=0.66\columnwidth]{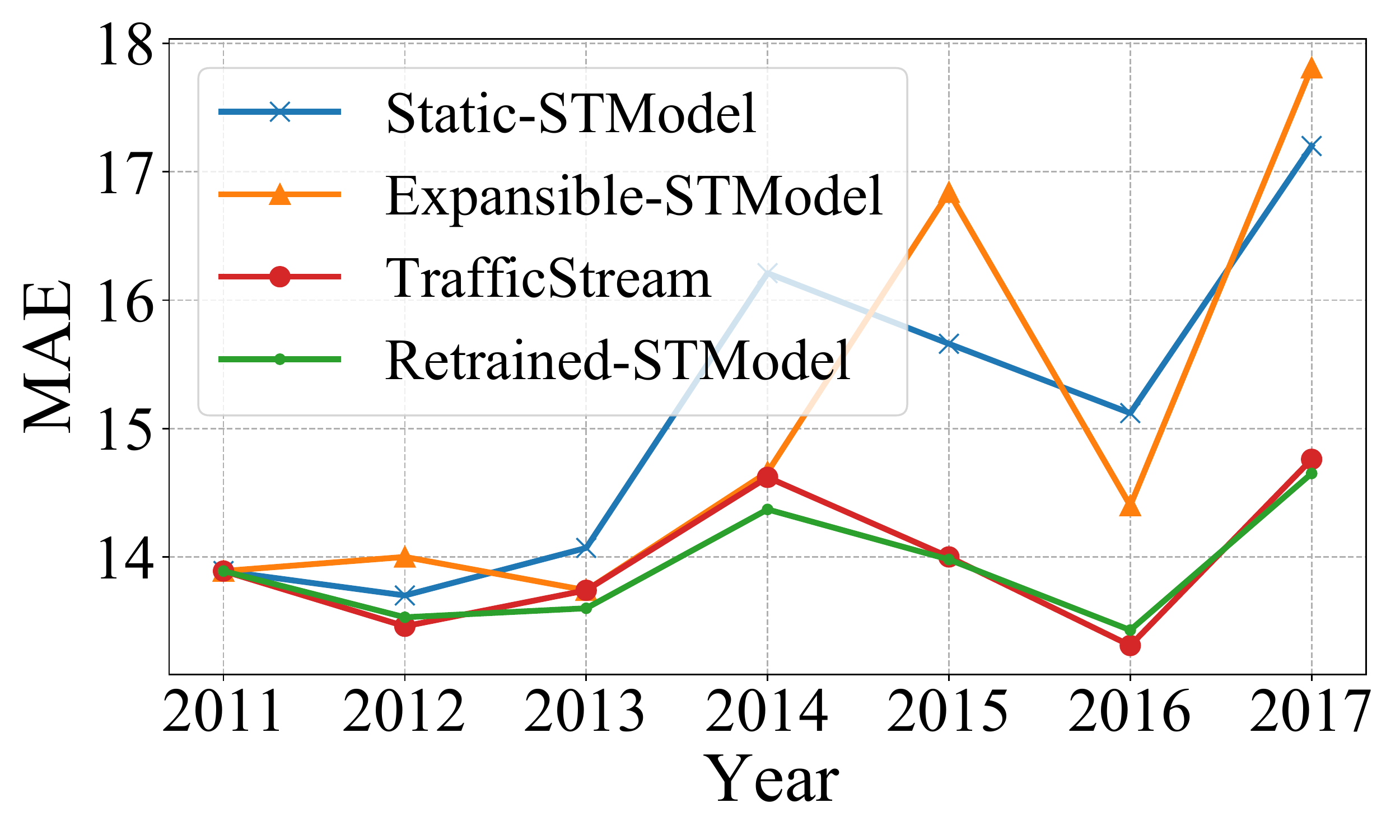}
    \label{fig:d}
}
\subfigure[30min RMSE]{
    \includegraphics[width=0.66\columnwidth]{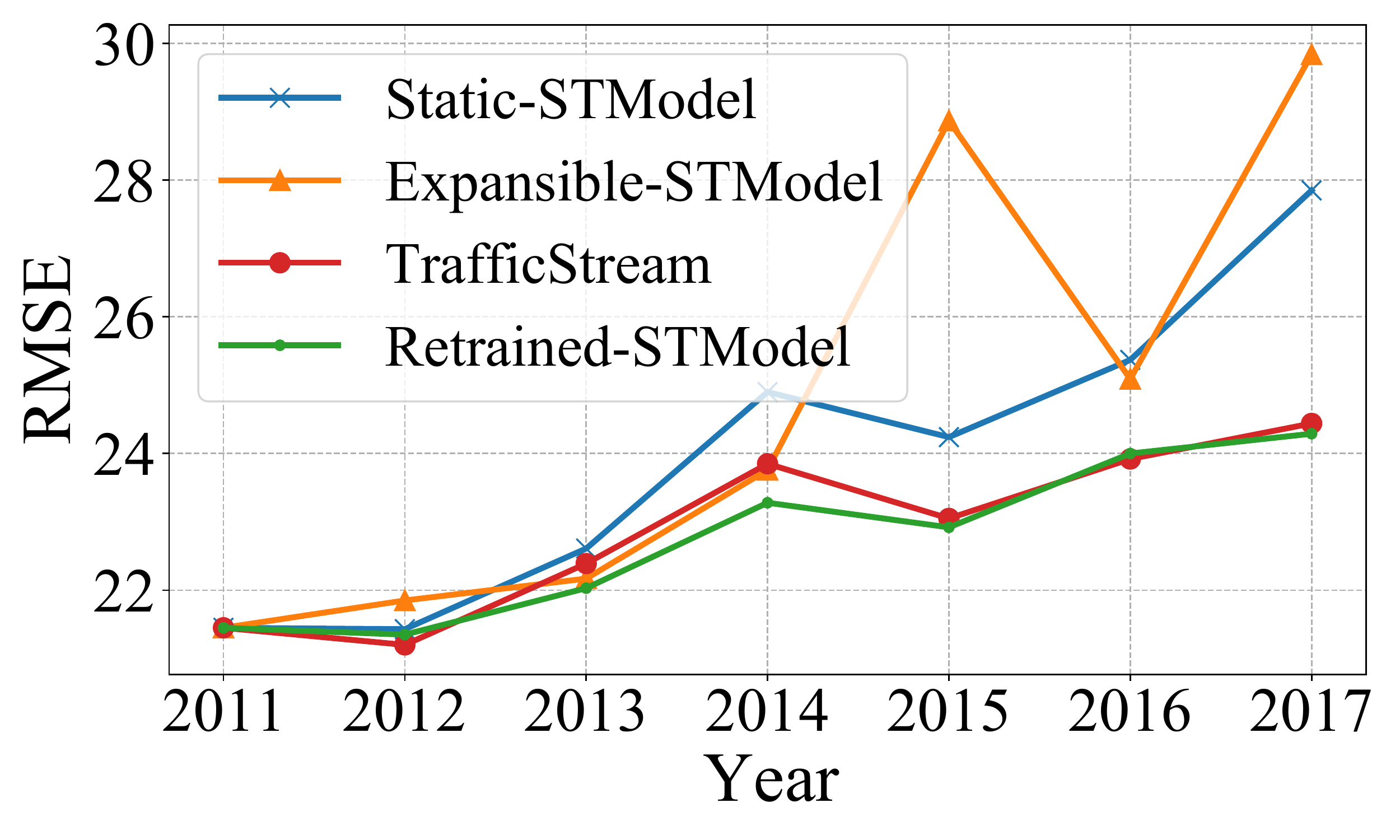}
    \label{fig:e}
}
\subfigure[Running time per epoch]{
    \includegraphics[width=0.66\columnwidth]{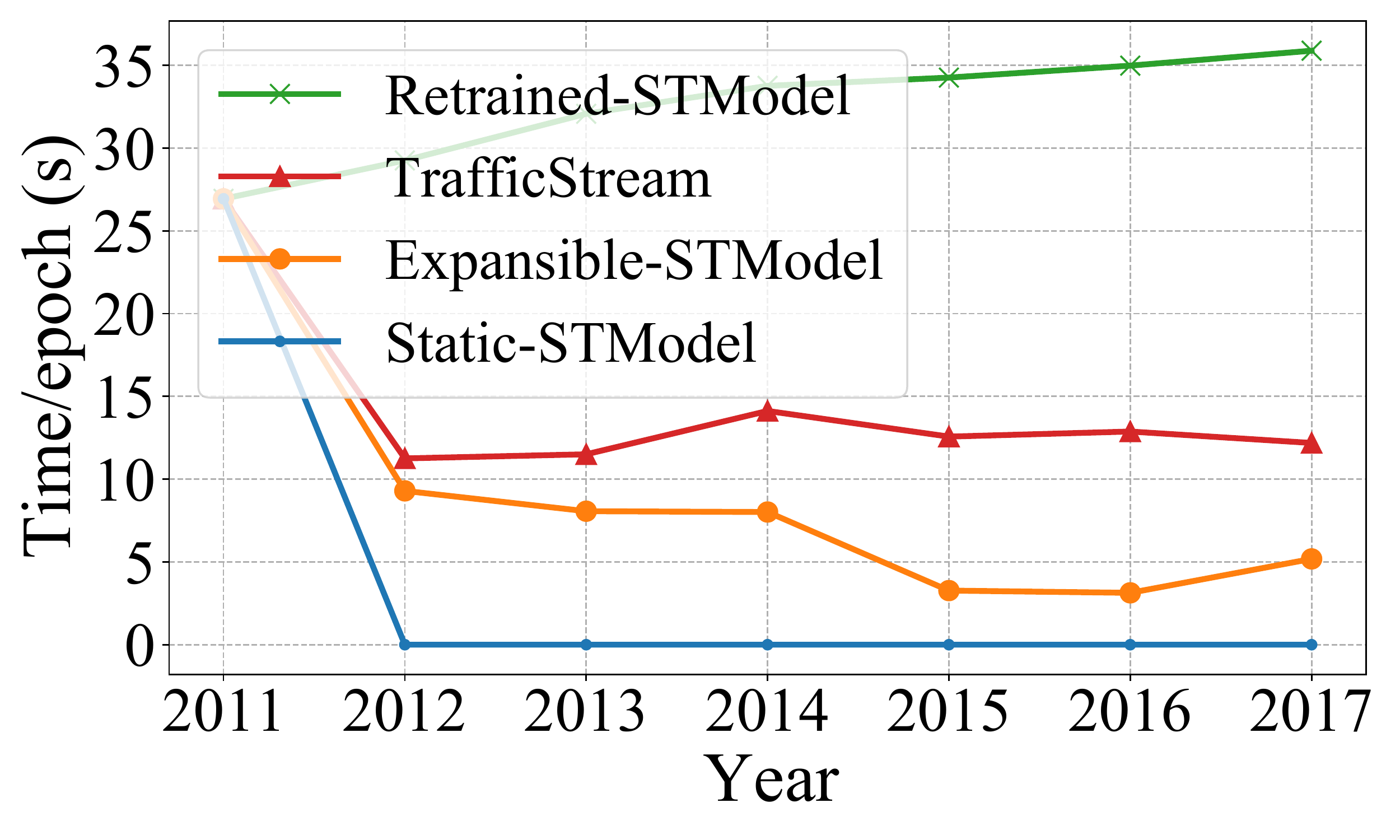}
    \label{fig:f}
}
\caption{MAE, RMSE and running time of traffic forecasting on consecutive years.}
\label{fig:res}
\end{figure*}


\subsection{Experimental Settings}
Data of each year is split into training, validation and testing sets with ratio 6:2:2 in the time dimension. Our task is to forecast traffic flows in the next hour with data from the last hour as $T=K=12$.
AdamW \cite{loshchilov2017decoupledadamw} is used as the optimizer with learning rate 0.005. Batch size is 128 and we train 100 epochs for each year with early stopping to accelerate training process. We apply random search strategy to find the best hyper-parameters for TrafficStream. 

To evaluate the performance of our framework, we adopt Mean Absolute Errors (MAE), Root Mean Squared Errors (RMSE) and Mean Absolute Percentage Errors (MAPE) as the metric for effectiveness. Total training time and training time per epoch are used to reveal model efficiency. 

\subsection{Baseline Methods}
\begin{itemize}
    \item{\textit{SVR}} \cite{svr}: Support Vector Regression adapts support vector machine for regression task.
    \item{\textit{GRU}} \cite{gru}: Gated Recurrent Unit is an RNN model that leverages gated mechanism.
    \item \textit{STGCN} \cite{stgcn}: Spatial-Temporal Graph Convolutional Network adopts graph convolution and gated CNN to extract spatial and temporal patterns. We train a STGCN using data from all nodes every year. 
    \item \textit{STSGCN} \cite{stsgcn}: Spatial-Temporal Synchronous Graph Convolutional Network synchronously captures the localized spatial-temporal correlations. We train a STSGCN using data from all nodes every year. 
    \item \textit{Retrained-STModel}: We train a surrogate model introduced in Sec \ref{sec:41} using data from all nodes every year. We use models at the last year as initialization.
    \item \textit{Static-STModel}: We utilize data at the first year (i.e., 2011) to train a surrogate model, and forecast traffic flow after 2011 using the surrogate model directly.
    \item \textit{Expansible-STModel}: We train a surrogate model in an online manner, that is, at every year, we only train the data from the newly added nodes with the models trained at the last year as initialization.
    \item \textit{TrafficStream}: Our proposed framework. We train a surrogate model in an online manner with new pattern fusion and historical knowledge consolidation. 
\end{itemize}

\subsection{Experimental Results}

In this section, we compare TrafficStream with other baselines on the traffic flow forecasting task and discuss the results from effectiveness and efficiency.

\subsubsection{Effectiveness} Table. \ref{tab:res} shows the overall performance of traffic forecasting at different time granularity (i.e., 15-, 30-, 60-minutes in the future). We compute the averaged MAE, RMSE and MAPE in 7 years of each model. Our model, TrafficStream achieves lower prediction error compared with the traditional online training methods (i.e., Static-STModel and Expansible-STModel). And it reaches almost the same effect as the upper bound (i.e., Retrained-STModel, STGCN and STSGCN). Besides, STModel adopts GCN and CNN to capture spatial-temporal correlation. It is significantly better than simple methods like SVR and LSTM, and has comparable performance with the advanced model like STGCN.

Figure \ref{fig:a}, \ref{fig:b}, \ref{fig:d} and \ref{fig:e} show the effect of models on consecutive years. Every year, as Static-STModel does not update the model incrementally, it cannot capture new patterns on the traffic network, resulting in a rapid increase in prediction errors. Expansible-STModel only trains incrementally on new nodes every year and does not maintain the patterns of other nodes, leading to the problem of catastrophic forgetting. The expressive ability of the model will become unsatisfactory and unstable. Especially when the patterns of new nodes are inconsistent with other nodes (e.g., 2015), Expansible-STModel will get very poor prediction results. TrafficStream can maintain good and stable forecasting, indicating that it can continuously learn new traffic patterns while maintaining existing knowledge.

\subsubsection{Efficiency} Table \ref{tab:res}, Figure \ref{fig:c} and Figure \ref{fig:f} show the running time of each model, including total training time and training time per epoch, and we compute the averaged time in 7 years. The training time of Retrained-STModel and STGCN is much higher than online methods (i.e., Expansible-STModel and TrafficStream), which is about 3 times longer. It proves that TrafficStream can maintain high efficiency while achieving accurate prediction, providing the opportunity for real-world applications. Note that Static-STModel uses the model trained in the first year to predict the annual traffic flows after that, so its training time is 0 after 2011.

\begin{figure}[t]
\centering
\includegraphics[width=0.9\columnwidth]{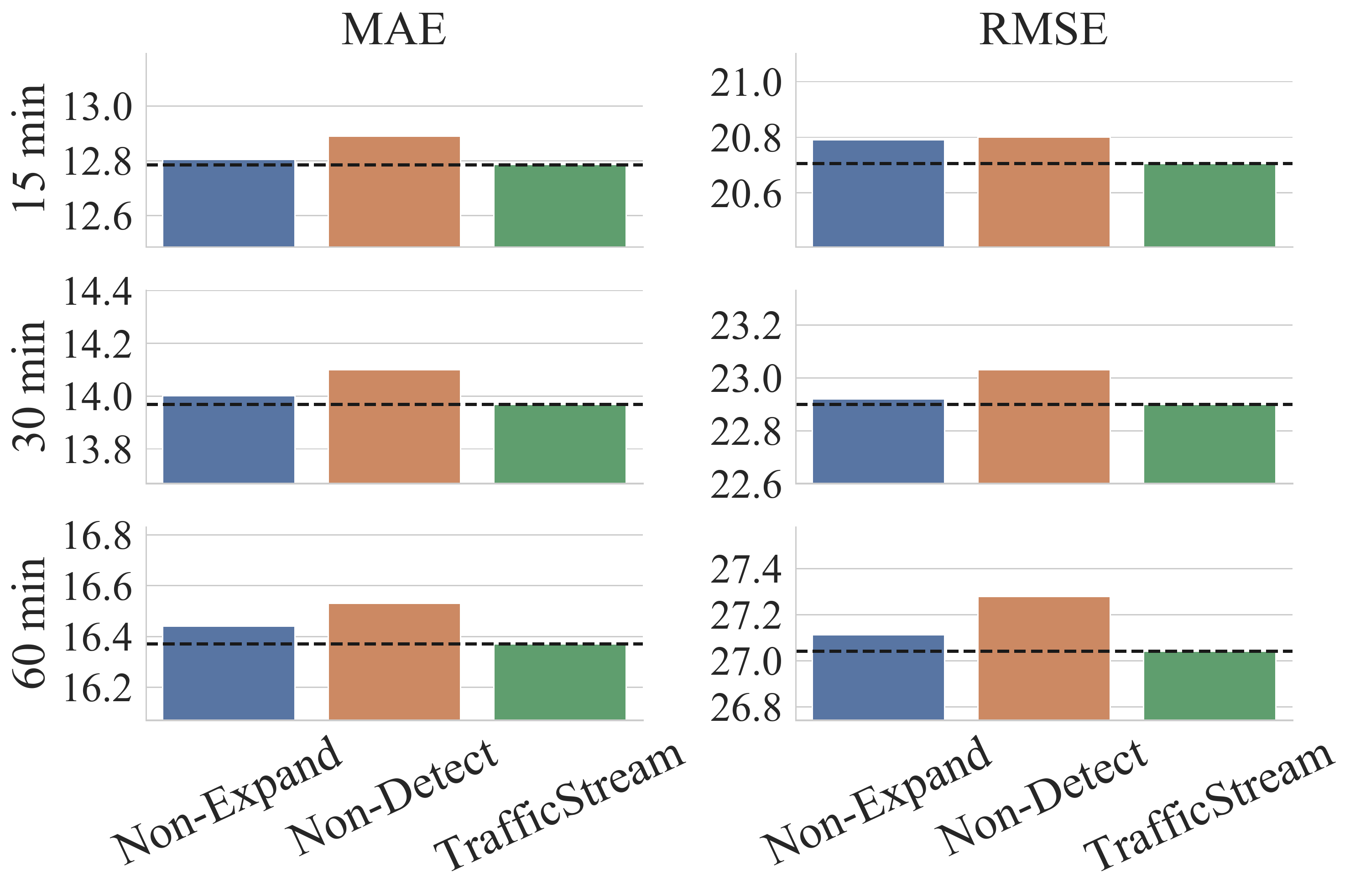}
\caption{Ablation study of new pattern fusion.}
\label{fig:new}
\end{figure}

\begin{figure}[t]
\centering
\includegraphics[width=0.9\columnwidth]{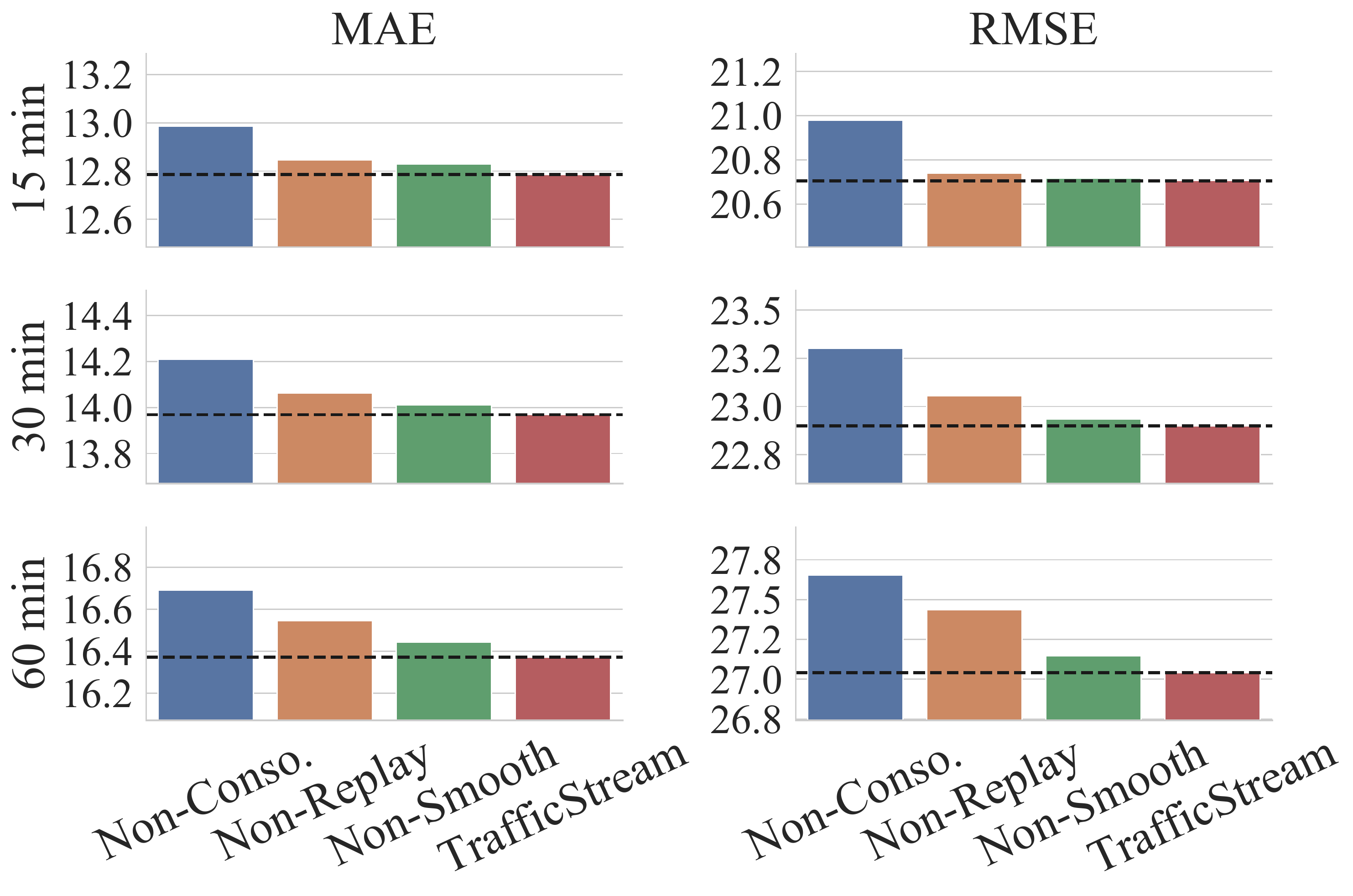}
\caption{Ablation study of historical knowledge consolidation.}
\label{fig:consolidation}
\end{figure}

\subsection{Ablation Study}

\subsubsection{Impact of New Pattern Fusion}

We first analyze the impact of the fusion of new patterns on the forecasting performance, as shown in Figure \ref{fig:new}. \textit{Non-Expand} refers to not integrating the patterns of new nodes and only integrating the new patterns of existing nodes. On the contrary, \textit{Non-Detect} represents only integrating the patterns of new nodes. \textit{TrafficStream} integrates all possible new patterns. It shows that it is not enough only to learn any parts of new patterns. Our model fuse all potential new patterns well so as to obtain the latest and best parameters to forecast the traffic flows in the future.

\subsubsection{Impact of Knowledge Consolidation}

We then analyze the influence of historical knowledge on the forecasting performance, as shown in Figure \ref{fig:consolidation}. \textit{Non-Conso.} represents that it does not consolidate any knowledge learned by the previous model at all. \textit{Non-Replay} and \textit{Non-Smooth} respectively refer to training without data replay and without model smoothing. \textit{TrafficStream} utilizes both approaches simultaneously.The results of not consolidating historical knowledge are the worst, indicating that the forgetting of historical knowledge occurs, so that the prediction of some nodes becomes worse. The two consolidation approaches can avoid the forgetting problem to a certain extent, and the approach of data replay is better. Besides, the combination of the two further improves the predictive ability of the model. 
\section{Conclusion}

In this paper, we propose a GNN framework for streaming traffic flow forecasting. 
Continual learning is introduced to achieve efficient update and effective prediction in streaming traffic data. 
The model is evaluated and proved to be very effective on PEMS3-Dynamic dataset. Compared with the traditional retraining approach, TrafficStream achieves high prediction accuracy while reducing training complexity greatly.
\clearpage

\bibliographystyle{named}
\bibliography{Sub/ref}

\end{document}